\documentclass{article}

\usepackage[nonatbib,preprint]{neurips_2024}
\usepackage{amsmath,amsfonts,bm}

\def\eqref#1{equation~\ref{#1}}
\def\1{\bm{1}}

\def\vtheta{{\bm{\theta}}}
\def\vepsilon{{\bm{\varepsilon}}}
\def\vsigma{{\bm{\sigma}}}
\def\vphi{{\bm{\phi}}}
\def\vxi{{\bm{\xi}}}

\def\vb{{\bm{b}}}

\def\vd{{\bm{d}}}

\def\vg{{\bm{g}}}

\def\vt{{\bm{t}}}
\def\vu{{\bm{u}}}

\def\vx{{\bm{x}}}

\def\evd{{d}}

\def\mC{{\bm{C}}}

\DeclareMathAlphabet{\mathsfit}{\encodingdefault}{\sfdefault}{m}{sl}
\SetMathAlphabet{\mathsfit}{bold}{\encodingdefault}{\sfdefault}{bx}{n}

\def\gE{{\mathcal{E}}}

\def\gL{{\mathcal{L}}}

\def\gN{{\mathcal{N}}}

\def\gU{{\mathcal{U}}}

\def\sI{{\mathbb{I}}}

\def\sR{{\mathbb{R}}}

\usepackage{times,cite,w-thm}
\usepackage[T1]{fontenc}
\usepackage[utf8]{inputenc}
\usepackage{w-greek}
\usepackage{amsmath, array}% for pmb
\usepackage{amssymb}
\usepackage{bm}
\usepackage{textcomp}
\usepackage{siunitx}
\usepackage{tikz}
\usepackage{graphicx}
\usetikzlibrary{positioning, shapes.geometric, arrows}
\usetikzlibrary{decorations.pathreplacing}
\usetikzlibrary{calc}
\usepackage{comment}
\usepackage{subcaption}
\tikzset{
    ncbar angle/.initial=90,
    ncbar/.style={
        to path=(\tikztostart)
        -- ($(\tikztostart)!#1!\pgfkeysvalueof{/tikz/ncbar angle}:(\tikztotarget)$)
        -- ($(\tikztotarget)!($(\tikztostart)!#1!\pgfkeysvalueof{/tikz/ncbar angle}:(\tikztotarget)$)!\pgfkeysvalueof{/tikz/ncbar angle}:(\tikztostart)$)
        -- (\tikztotarget)
    },
    ncbar/.default=0.5cm,
}
\tikzset{square left brace/.style={ncbar=0.3cm}}
\tikzset{square right brace/.style={ncbar=-0.3cm}}

\usepackage[colorlinks, allcolors=blue]{hyperref}
\usepackage[noabbrev]{cleveref}
\usepackage{orcidlink}

\usepackage[normalem]{ulem}
\usepackage{color}

\newcommand{\FEsq}{FE$^2\,$} %FE-2
\newcommand{\tr}{\mathrm{tr}\,}

\def\macStrain{{\hat{\bm{\varepsilon}}}} % strain tensor at macroscale integration point j 
\def\macStress{{\hat{\bm{\sigma}}}} % stress tensor at macroscale integration point j

\graphicspath{{Figs/}}

\newlength{\mywidth}
\title{Enhancing Multiscale Simulations with Constitutive Relations-Aware Deep Operator Networks}

\newcommand{\orcidA}{\orcidlink{0000-0003-3650-4107}}
\newcommand{\orcidB}{\orcidlink{0000-0002-4999-4558}}
\newcommand{\orcidC}{\orcidlink{0000-0003-1849-0784}}
\newcommand{\orcidD}{\orcidlink{0000-0002-6850-6409}}

\author{%
\orcidA~Hamidreza Eivazi$^{\,\dagger,}$\thanks{Corresponding author. $\dagger$ These authors contributed equally to this work.} \\
\small{Institute for Software and Systems Engineering} \\
\small{Clausthal University of Technology} \\
\small{38678 Clausthal-Zellerfeld, Germany} \\
\small{\texttt{he76@tu-clausthal.de}}
\And
Mahyar Alikhani$^{\,\dagger}$ \\
\small{Institute for Software and Systems Engineering} \\
\small{Clausthal University of Technology}\\
\small{38678 Clausthal-Zellerfeld, Germany} \\
\small{\texttt{mahyar.alikhani@tu-clausthal.de}}
\And
\orcidB~Jendrik-Alexander Tröger\\
\small{Institute of Applied Mechanics} \\
\small{Clausthal University of Technology} \\
\small{38678 Clausthal-Zellerfeld, Germany} \\
\small{\texttt{jendrik-alexander.troeger@tu-clausthal.de}}
\And
Stefan Wittek\\
\small{Institute for Software and Systems Engineering} \\
\small{Clausthal University of Technology} \\
\small{38678 Clausthal-Zellerfeld, Germany} \\
\small{\texttt{stefan.wittek@tu-clausthal.de}}
\And
\orcidC~Stefan Hartmann\\
\small{Institute of Applied Mechanics} \\
\small{Clausthal University of Technology} \\
\small{38678 Clausthal-Zellerfeld, Germany} \\
\small{\texttt{stefan.hartmann@tu-clausthal.de}}
\And
\orcidD~Andreas Rausch\\
\small{Institute for Software and Systems Engineering} \\
\small{Clausthal University of Technology} \\
\small{38678 Clausthal-Zellerfeld, Germany} \\
\small{\texttt{andreas.rausch@tu-clausthal.de}}
}

\begin{document}
\hypersetup{
  pdftitle={Enhancing Multiscale Simulations with Constitutive Relations-Aware Deep Operator Networks},
  pdfsubject={},
  urlcolor=magenta,
  citecolor=blue,
  linkcolor=black
  }

\maketitle     

\begin{abstract}
Multiscale problems are widely observed across diverse domains in physics and engineering. Translating these problems into numerical simulations and solving them using numerical schemes, e.g. the finite element method, is costly due to the demand of solving initial boundary-value problems at multiple scales. On the other hand, multiscale finite element computations are commended for their ability to integrate micro-structural properties into macroscopic computational analyses using homogenization techniques. Recently, neural operator-based surrogate models have shown trustworthy performance for solving a wide range of partial differential equations. In this work, we propose a hybrid method in which we utilize deep operator networks for surrogate modeling of the microscale physics. This allows us to embed the constitutive relations of the microscale into the model architecture and to predict microscale strains and stresses based on the prescribed macroscale strain inputs. Furthermore, numerical homogenization is carried out to obtain the macroscale quantities of interest. We apply the proposed approach to quasi-static problems of solid mechanics. The results demonstrate that our constitutive relations-aware DeepONet can yield accurate solutions even when being confronted with a restricted dataset during model development.
\end{abstract}
%% maketitle must follow the abstract.
              % Produces the title.

%%%%%%%%%%%%%%%%%%%%%%%%%%%%%%
\section{Introduction}

\paragraph{Multiscale computations} Multiscale problems in physics involve phenomena that occur across multiple length or time scales. These problems often exhibit complex interactions and behaviors at different scales, making them challenging to analyze and simulate directly with conventional methods. Many physical systems and phenomena are described by partial differential equations (PDEs), which are mathematical equations that relate various properties across space and time. PDEs capture fundamental principles governing the behavior of continuous physical systems, but solving them directly for multiscale problems can be computationally demanding and impractical due to the need for fine resolution across all scales of interest. In multiscale computational mechanics, the so-called \FEsq approach \cite{smitbrekelmansmeijer1998,feyel1999,MieheKoch2002,KouznetsovaGeersBrekelmans2004,HARTMANN2023} tackles challenges posed by heterogeneous materials and complex microstructures at varying scales. The PDEs to be solved in solid mechanics are the local balance of linear momentum on macro- and microscale. The \FEsq approach integrates macroscale and microscale simulations, typically by employing finite element analysis at both scales to capture macro- and microscale behaviors. However, such methods can be computationally intensive due to the need for extensive microscale simulations throughout the simulation of the macroscale domain. To tackle these obstacles, integrating machine learning (ML)-based approaches into multiscale simulations presents promising benefits, particularly in accelerating the bridging between different length scales within the domain of interest, see, for instance, \cite{fengzhangkhandelwal2022,Kalina2023,mca28040091,Trger2023,KALINA2024}. Although purely data-driven surrogate models have shown promising results in this field, they often require a large amount of data and lack both robustness and generalizability.

\paragraph{Operator learning} Recently, a new branch of ML research, the so-called operator learning, has made substantial advances for solving PDEs by providing methods for learning operators. In this context, operators are understood as maps between infinite-dimensional spaces, which is in contrast to functions that are maps between finite-dimensional vector spaces \cite{kovachki2023neural}. Neural operators have been shown as a trustworthy replacement for numerical approaches for solving PDEs, since they offer enhanced capabilities in capturing complex patterns and relationships within data. According to \cite{kovachki2023neural}, neural operators are known as a class of models that guarantee both discretization-invariance and universal approximation. In contrast to classical physics-informed neural networks \cite{raissi2019physics}, operator networks can obtain a solution for a new instance of the PDE only in a forward pass and do not require additional training. Successful operator learning approaches reported in current literature are, among others, deep operator networks (DeepONets) \cite{lu2021learning} and its proper orthogonal decomposition (POD)-based extension (POD-DeepONet) \cite{lu2022poddeeponet}, Fourier neural operator (FNO) \cite{li2020fourier}, and PCA-based neural networks (PCANN) \cite{bhattacharya2020model}. In this article, we focus on DeepONet and its extensions. More recently, advanced architectures equipped with attention mechanism \cite{zhao2023pinnsformer} or diffusion modules \cite{lippe2023pderefiner} have been proposed to approximate solution operator of PDEs. Among all proposed frameworks for operator learning, DeepONet stands out for its innovative approach and adaptability across various applications, as proposed by \cite{lu2022poddeeponet, Lin_2021, CAI2021110296, dileoni2021deeponet, MAO2021110698,eivazi2024nonlinear}. DeepONets have recently been introduced for applications in computational solid mechanics, see \cite{HE2023116277,HE2024107258}, however, not in a multiscale context.

\paragraph{Our contribution} In this contribution, we introduce a flexible and constitutive relations-aware learning-based PDE surrogate for predicting microscale physics based on operator networks. Our work presents the first operator learning-based surrogate model that provides microscale solutions for the representative volume element (RVE), which represents the microstructure under consideration. Thus, our approach enables the incorporation of known microscale physical relations into the model. This contrasts with the majority of the literature, which has predominantly focused on learning mappings substituting the entire microscale boundary-value problem. We showcase the applicability of our framework for multiscale \FEsq computations, facilitating the incorporation of micro-mechanical material structures into macroscale simulations. Our approach leverages deep operator networks following the work by \cite{lu2021learning} and formulates a constitutive relations-aware model. Subsequently, homogenized quantities are computed similarly to traditional \FEsq computations, and the consistent tangent matrix is derived using automatic differentiation. We apply the proposed hybrid approach to quasi-static problems in solid mechanics. The results highlight the method's ability to produce accurate solutions, even with a limited dataset during model development. Additionally, we demonstrate that an efficient implementation through state-of-the-art high-performance computing libraries and just-in-time compilation can yield multiple orders of magnitude in speed-up compared to conventional \FEsq schemes.

%%%%%%%%%%%%%%%%%%%%%%%%%%%%%%
\section{Methodology}

\begin{figure}[ht]
\begin{center}
\scalebox{0.6}{
\begin{tikzpicture}[->,>=stealth',shorten >=1pt,auto,node distance=5cm,
                      thick,main node/.style={font=\Large\bfseries}]
    % Macroscale Image
    \node (macroscale) at (0,0) {\includegraphics[width=0.35\mywidth]{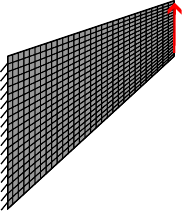}};
    \node[above=1mm of macroscale,font=\Large\bfseries] (macroscale_label) {Macroscale};
    \node (1) at (1, 0) {};

    % Element
    \node[right=.5cm of macroscale] (element) {\includegraphics[width=2cm]{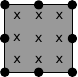}};
    \node[above=2mm of element,font=\Large\bfseries] {};
    \node[font=\Large\bfseries] (j) at (3.6, 0) {$j$};

    \path[every node/.style={}]
      (1) edge node [above] {element} (element);

    % Operator Network Box
    \node[draw, rectangle, minimum width=7cm, minimum height=4cm, below=2cm of element] (opnet) {};
    % \node[above=0mm of opnet.north] (opnet_label) {Operator Network};
    \node at (opnet) {\includegraphics[width=2cm]{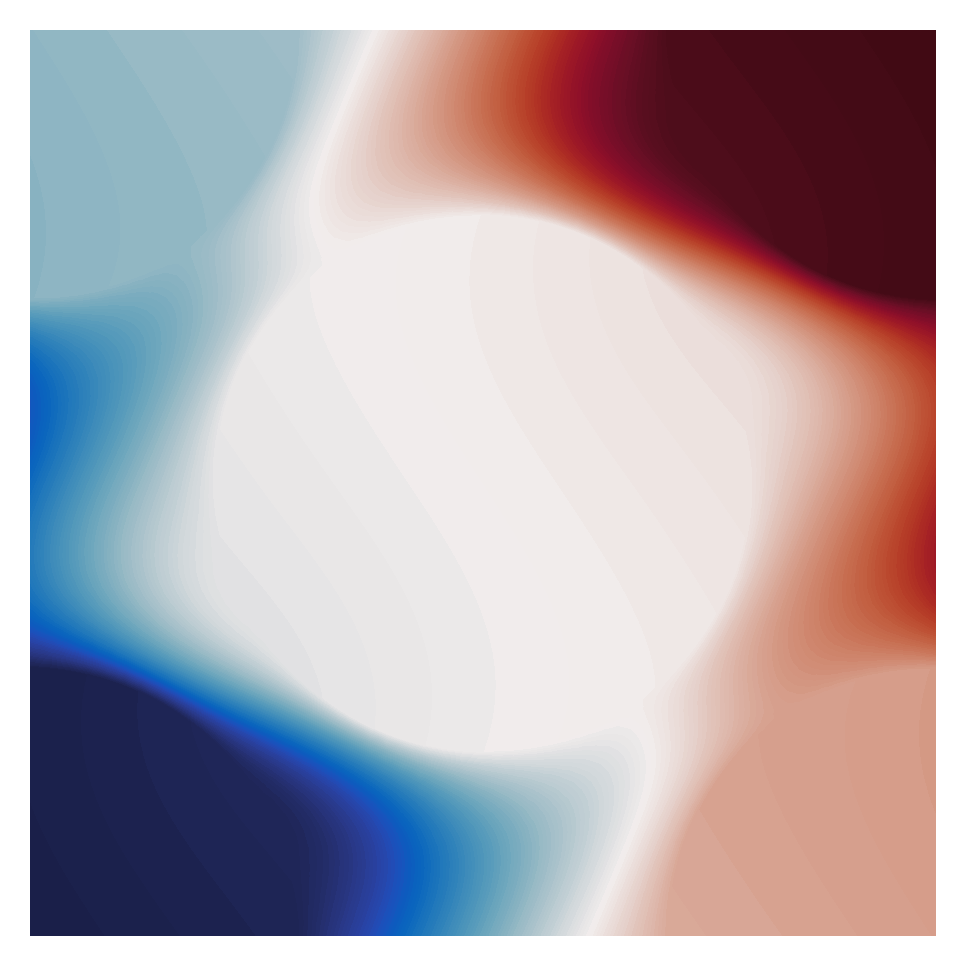}
                    \includegraphics[width=2cm]{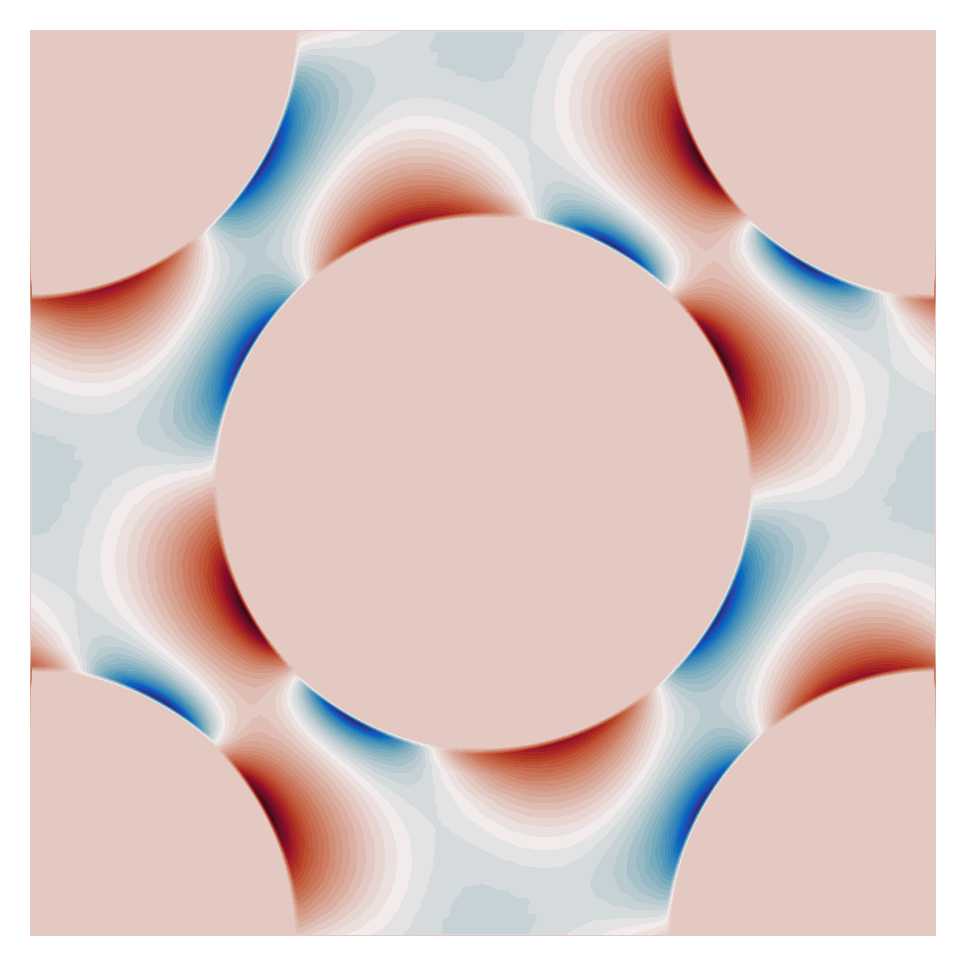}
                    \includegraphics[width=2cm]{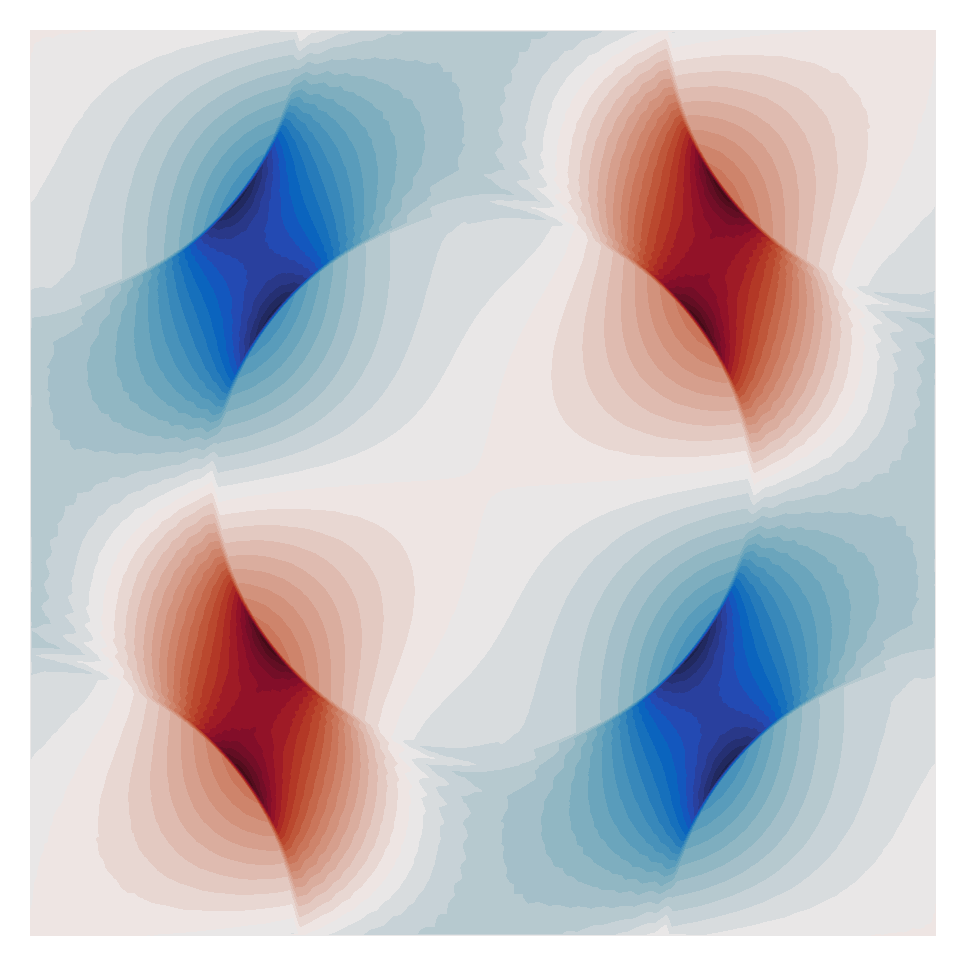}
                    };
    \node[above=-6mm of opnet,align=center,font=\large\bfseries] {Microscale};
    \node[align=center,font=\large] (mapu) at (1.75, -4.0) {$\macStrain \to \vu$};
    \node[align=center,font=\large] (mape) at (3.85, -4.0) {$\vu \to \vepsilon$};
    \node[align=center,font=\large] (maps) at (6.0, -4.0) {$\vepsilon \to \vsigma$};
    
    \node[align=center] (u) at (1.75, -6.5) {Operator\\ network};
    \node[right=1.5em of u, align=center] (epsilon) {Kinematic\\ Relation};
    \node[right=0.8em of epsilon, align=center] {Constitutive\\ Relation};

    \begin{scope}[transform canvas={xshift=-.5em}]
    \tikzstyle{every to}=[draw]
    \draw (element) to[style={font=\LARGE\bfseries}] node[left] {$(\macStrain^j, \vxi)$} (opnet);
    \end{scope}

    \begin{scope}[transform canvas={xshift=.5em}]
    \tikzstyle{every to}=[draw]
    \draw (opnet) to[style={font=\LARGE\bfseries}] node[right] {$(\macStress^j, \mC^j)$} (element);
    \end{scope}

    \node[left=10mm of opnet, align=center] (rve) {\includegraphics[width=0.15\mywidth]{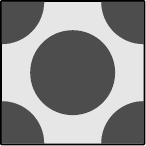}};
    \node[below=0.1em of rve, align=center,font=\Large\bfseries] {RVE};

    \node[right=6cm of element, align=center] (trunk) {\includegraphics[width=0.25\mywidth]{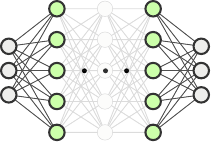}};
    \node[above=0.1cm of trunk, align=center] (branch) {\includegraphics[width=0.25\mywidth]{nn.pdf}};

    \node[left=1em of branch, align=center,font=\LARGE\bfseries] {$\macStrain$};
    \node[left=1em of trunk, align=center,font=\LARGE\bfseries] {$\vx$};

    \node[above=-2mm of branch, align=center] {Branch};
    \node[above=-2mm of trunk, align=center] (trunk_title) {Trunk};

    \node[draw, circle,minimum height=0.5cm, right=1.5cm of trunk_title] (dot) {$\times$};

    \path[every node/.style={}]
      (branch) edge node [above] {$\vb$} (dot);

    \path[every node/.style={}]
      (trunk) edge node [below] {$\vt$} (dot);

    \node[right=1em of dot, align=center,font=\LARGE\bfseries] (output) {$\vu(\vx, \macStrain)$};

    \path[every node/.style={}]
      (dot) edge node [right] {} (output);

    \node[draw, rectangle, minimum width=0.25\mywidth, minimum height=1.7cm, below=4cm of trunk,font=\LARGE\bfseries] (trunk2) {${\vphi}(\vx)$};
    \node[above=0.3cm of trunk2, align=center] (branch2) {\includegraphics[width=0.25\mywidth]{nn.pdf}};

    \node[left=1em of branch2, align=center,font=\LARGE\bfseries] {$\macStrain$};
    \node[left=1em of trunk2, align=center,font=\LARGE\bfseries] {$\vx$};

    \node[above=-2mm of branch2, align=center] {Branch};
    \node[above=-5.5mm of trunk2, align=center] (trunk_title2) {POD basis};

    \node[draw, circle,minimum height=0.5cm, right=1.5cm of trunk_title2] (dot2) {$\times$};

    \path[every node/.style={}]
      (branch2) edge node [above] {$\vb$} (dot2);

    \path[every node/.style={}]
      (trunk2) edge node [below] {$\vphi$} (dot2);

    \node[right=1em of dot2, align=center,font=\LARGE\bfseries] (output2) {$\vu(\vx, \macStrain)$};

    \path[every node/.style={}]
      (dot2) edge node [right] {} (output2);

    \node[left=3em of macroscale_label,font=\Large\bfseries] (a) {(a)};
    \node[below=5.5cm of a,font=\Large\bfseries] (b) {(b)};

    \node[right=11cm of a,font=\Large\bfseries] (c) {(c)};
    \node[right=11cm of b,font=\Large\bfseries] {(d)};
    
\end{tikzpicture}}
\end{center}
\caption{A schematic view of the proposed method. (a) representation of the hybrid multiscale method. (b) the representative volume element (RVE) employed in this study; light gray indicates the matrix and dark gray shows the fiber. (c) and (d) depict DeepONet and POD-DeepONet architectures, respectively.}
\label{fig:fedeeponet}
\end{figure}

Through a multiscale approach, the constitutive behavior observed at the microscale can be applied to macroscale analysis. This involves employing finite element to discretize a heterogeneous microscale domain represented by an RVE, see, for example, \cite{FEYEL20033233}. 
In this context, by applying the macroscale strain $\macStrain^{\,j} \in \mathbb{R}^3$ (e.g. for a two-dimensional setup) of any macroscale integration point $j$ as a boundary condition for the RVE, we can capture the deformation behavior of the RVE at its boundaries. The acquired microscale boundary-value problem can be solved using the proposed operator network surrogate model.~\Cref{fig:fedeeponet}(a) depicts a schematic view of the hybrid method.
Our focus here is limited to periodic displacement boundary conditions at the microscale. Let us consider the microscale domain of interest $\Omega$ to be a symmetric and zero-centered unit cell such that
\begin{equation}
    \Omega=\left\{\vx=\left\{x^d\right\} \in \mathbb{R}^d \left\lvert\, \frac{-L}{2} \leq x^d \leq \frac{L}{2}\right.\right\},
\end{equation}
for any point $\vx$ where $L$ indicates the edge length of the domain $\Omega$, and $d$ the dimension of the problem under consideration. We develop our framework by letting the operator network output the displacement vector $\vu(\vx, \macStrain)$ at any point $\vx \in \Omega$ for a given prescribed macroscale strain $\macStrain$. Note that we omit the superscript $j$ in the following for brevity. Our goal is to learn the solution operator of the microstructure mechanics for different periodic boundary conditions imposed by the prescribed macroscale strain $\macStrain$.
%, i.e.\ we aim at a parameterization of the solution operator in the boundary conditions. 
To this end, we consider the operator network $\gN_{\vtheta} $ parameterized by $\vtheta$ such that
\begin{equation}
    \vu(\vx, \macStrain) \approx \gN_{\vtheta}(\vx, \macStrain; \vtheta).
\end{equation}
Further, we embed well-known physical relations by utilizing the linearized kinematic relation 
\begin{equation}
    \label{eq:kinematic}
    \vepsilon(\vx, \macStrain) = \dfrac{1}{2} (\nabla \vu(\vx, \macStrain) + \nabla \vu^T(\vx, \macStrain)), \quad \vx \in \Omega,
\end{equation}
mapping the microscale displacement vector field $\vu(\vx, \macStrain)$ to the strain tensor field $\vepsilon(\vx, \macStrain)$. Additionally, constitutive relations are employed to describe the dependence of the stress tensor $\vsigma(\vx,\macStrain)$ on the strain tensor field $\vepsilon(\vx,\macStrain)$
\begin{equation}
    \label{eq:constRelation}
    \vsigma(\vx,\macStrain) = \vg(\vepsilon(\vx, \macStrain)).
\end{equation}
Note that we restrict ourselves to elastic material behavior in the present study, i.e.\ $\vg$ can represent any linear or nonlinear constitutive elasticity relation.
As a specific example of a heterogeneous microstructure, we use a fiber-reinforced material described in \cref{sec:problem_setup}, in which a nonlinear elastic material behavior is assumed for the matrix and linear elasticity for the fibers. To obtain the sought macroscale quantities -- macroscale stress $\macStress$ -- we draw on common numerical homogenization, see, for example, \cite{MieheKoch2002} for further details. Further, the consistent tangent matrix $\mC$ is obtained using automatic differentiation. 

\subsection{Deep Operator Networks (DeepONets)}

In the following, we employ POD-DeepONet \cite{lu2022poddeeponet} to develop our PDE-based surrogate model for the microscale. To this end, a brief overview of DeepONet and its POD-based version is recapped first.

The extension of the universal approximation theorem to encompass the approximation of nonlinear operators, as discussed in \cite{392253}, confirms DeepONet as a powerful tool for solving PDEs. A DeepONet represents a deep-learning structure designed to approximate a nonlinear operator on a discretized input field or function at any chosen query point. This method is particularly useful for training models that can solve sets of PDEs, allowing to query solutions at various spatial and temporal coordinates. 
We consider a stacked DeepONet with bias following the work of \cite{lu2021learning}. Let us consider $\gE$ and $\gU$ as two separable infinite-dimensional function spaces over bounded domains and assume that $\gN: \gE \to \gU$ is an arbitrary linear or nonlinear operator. A DeepONet approximates $\gN$ using two sub-networks, i.e. a trunk and a branch network, see \Cref{fig:fedeeponet}(c).
In our problem setup, the trunk network takes the coordinates $\vx$ as the input and outputs a set of basis functions. The branch network takes the prescribed macroscale strains $\macStrain$ as input and outputs the coefficients of the basis functions predicted by the trunk network. The operator $\gN$, that maps the input $\macStrain$ to the output function $\vu(\vx, \macStrain)$, can be approximated by linear reconstruction of the output function
\begin{equation}
   \gN(\macStrain)(\vx)\approx \sum_{k=1}^{p} b_k(\macStrain) t_k(\vx) + b_0,
\end{equation}
for any point $\vx$ in $\Omega$, where $b_0 \in \mathbb{R}$ indicates a bias, $\{b_1, b_2, \cdots, b_p\}$ are the $p$ outputs of the branch network, and $\{t_1, t_2, \cdots, t_p\}$ are the $p$ outputs of the trunk network representing the basis functions. The trunk net automatically learns this set of bases for the output function $\vu$ from the training data. 
In POD-DeepONet \cite{lu2022poddeeponet}, the trunk network is replaced by a set of POD bases, and the branch network learns their coefficients as depicted in \Cref{fig:fedeeponet}(d).
% , see \Cref{fig:architectureDeepONet}.
Thus, the output can be written as
\begin{equation}
   \gN(\macStrain)(\vx)\approx \sum_{k=1}^{p} b_k(\macStrain) \phi_k(\vx) + \phi_0(\vx),
\end{equation}
where $\{\phi_1, \phi_2, \cdots, \phi_p\}$ are the POD bases of the output function and $\phi_0$ is the so-called mean function. $\phi(\vx)$ can be obtained using interpolation for any point $\vx$ in the domain $\Omega$, \cite{bhattacharya2020model}. It has been shown that POD-DeepONets outperform DeepONets in most of the benchmark test cases for operator learning \cite{lu2022poddeeponet}. A DeepONet or POD-DeepONet can be trained by minimizing a loss function as
\begin{equation}
    \label{eq:loss_s}
    \gL(\vtheta) = \dfrac{1}{Nm}\sum_{i=1}^N \sum_{j=1}^{m} \left | \gN_{\vtheta}(\macStrain_i)(\vx_j) - \gN(\macStrain_i)(\vx_j)\right |^2,
\end{equation}
where $\gN_{\vtheta}$ denotes the deep operator network and $\vtheta$ is the collection of all trainable weight and bias parameters in the model. $N$ indicates the number of input samples for macroscale strains $\macStrain$ sampled from $\gE$, and $m$ is the number of collocation points in the domain $\Omega$.

\subsection{POD basis functions and Galerkin projection}

POD-DeepONet requires a set of precomputed POD basis functions or the so-called modes. These modes can be obtained from limited examples of the output function on a set of collocation points $\vxi \in \sR^m$ in domain $\Omega$ as $[\vu(\vxi, \macStrain^1),\ldots,\vu(\vxi, \macStrain^N)]$, which represents the snapshot matrix $\mathbf{U}$. The branch network maps the prescribed macroscale strain $\macStrain$ to the coefficients associated with these POD basis functions. We obtain the POD basis functions by applying singular-value decomposition (SVD) \cite{Sirovich_1987} directly on the snapshot matrix $\mathbf{U}$ after removing the mean,
\begin{equation}
    \scalebox{0.8}{
    \begin{tikzpicture}

        %%%%%%%%%%% Snapshot matrix of displacement at %%%%%%%%%%%
        \draw[black, thin, fill=gray!10] (-2.3,1) rectangle (-0.5, 5);

        % \draw  [black, thick, dashed, rounded corners] (0.2, 3.05) rectangle (0.7, 4.9);
        \draw [black, very thin] (-2, 4.3) -- (-2, 4.8);
        \node at (-2, 4) {\(u_{x}^{1}\)};
        \draw [black, very thin] (-2, 3.2) -- (-2, 3.7);

        % \draw  [black, thick, dashed, rounded corners] (0.2, 1.1) rectangle (0.7, 2.95);
        \draw [black, very thin] (-2, 1.25) -- (-2, 01.75);
        \node at (-2, 2) {\(u_{y}^{1}\)};
        \draw [black, very thin] (-2, 2.3) -- (-2, 2.8);

        \node at (-1.4, 4) {\ldots};        
        \node at (-1.4, 2) {\ldots};
        
        \draw [black, very thin] (-0.8, 4.3) -- (-0.8, 4.8);
        \node at (-0.8, 4) {\(u_{x}^{N}\)};
        \draw [black, very thin] (-0.8, 3.2) -- (-0.8, 3.7);
        
        \draw [black, very thin] (-.8, 1.25) -- (-0.8, 01.75);
        \node at (-0.8, 2) {\(u_{y}^{N}\)};
        \draw [black, very thin] (-0.8, 2.3) -- (-0.8, 2.8);

        \node at (-1.4, 0.5) {\Large{\(\mathbf{U}\)}};
        \node at (2.0, 0.5) {\Large{\(\mathbf{\Phi}\)}};
        %%%%%%%%%%% = %%%%%%%%%%%
        \node at (-0.25, 3){=};

        %%%%%%%%% square matrrix - left singular matrix- modes of displacement %%%%%%%%%
        \draw[black, draw=none, fill=gray!10] (0,1)  rectangle (4,5);
        \draw [black, thick] (0.3,1) to [square left brace ] (0.3,5);
        \draw [black, thick] (3.7,1) to [square right brace] (3.7,5);

        \draw  [black, thick, dashed, rounded corners] (0.2, 3.05) rectangle (0.7, 4.9);
        \draw [black, very thin] (0.45, 4.3) -- (0.45, 4.8);
        \node at (0.45, 4) {\(\phi_{x}^{1}\)};
        \draw [black, very thin] (0.45, 3.2) -- (0.45, 3.7);

        \draw  [black, thick, dashed, rounded corners] (0.2, 1.1) rectangle (0.7, 2.95);
        \draw [black, very thin] (0.45, 1.25) -- (0.45, 01.75);
        \node at (0.45, 2) {\(\phi_{y}^{1}\)};
        \draw [black, very thin] (0.45, 2.3) -- (0.45, 2.8);
        
        \draw  [black, thick, dashed, rounded corners] (0.9, 3.05) rectangle (1.4, 4.9);
        \draw [black, very thin] (1.15, 4.3) -- (1.15, 4.8);
        \node at (1.15, 4) {\(\phi_{x}^{2}\)};
        \draw [black, very thin] (1.15, 3.2) -- (1.15, 3.7);

        \draw  [black, thick, dashed, rounded corners] (0.9, 1.1) rectangle (1.4, 2.95);
        \draw [black, very thin] (1.15, 1.25) -- (1.15, 01.75);
        \node at (1.15, 2) {\(\phi_{y}^{2}\)};
        \draw [black, very thin] (1.15, 2.3) -- (1.15, 2.8);

        \node at (2.5, 2) {\textbf{\ldots}};
        \node at (2.5, 4) {\textbf{\ldots}};

        \draw  [black, thick, dashed, rounded corners] (3.3, 3.05) rectangle (3.8, 4.9);
        \draw [black, very thin] (3.55, 4.3) -- (3.55, 4.8);
        \node at (3.55, 4) {\(\phi_{x}^{N}\)};
        \draw [black, very thin] (3.55, 3.2) -- (3.55, 3.7);

        \draw  [black, thick, dashed, rounded corners] (3.3, 1.1) rectangle (3.8, 2.95);
        \draw [black, very thin] (3.55, 1.25) -- (3.55, 01.75);
        \node at (3.55, 2) {\(\phi_{y}^{N}\)};
        \draw [black, very thin] (3.555, 2.3) -- (3.555, 2.8);
      
        %%%%%%%%%% Matrix Sigma %%%%%%%%%%

      \draw[black, very thin, fill=gray!1] (4.2, 1) rectangle (6,5);
      \node at (5.1, 0.5){\large{$\mathbf{\Sigma}$}};
      \draw[line width=4, opacity=0.15, cap=round] (4.25,4.95) -- (5.95,2.95);
      \draw[very thin, dashed] (4.3, 1.1) rectangle (5.9, 2.7);
      \draw[black, thin, <->] (4.3, 2.8) -- (5.9, 2.8); 
      \node at (5, 3){$N$};
      
      %%%%%%%%%%%% % MAtrix V %%%%%%%%%%%% 
      \draw[black, thin, fill= gray!7] (6.2, 3.2) rectangle (8,5);
      \draw[black, thin, <->] (6.2, 3.05) -- (8, 3.05); 
      \draw[black, thin, <->] (8.15, 3.2) -- (8.15, 5); 
      \node at (7.4, 0.5) {\Large{$\mathbf{V^{*}}$}};
      \node at (8.4, 4.1){$N$};
      \node at (7.1, 2.8){$N$};
      
      % \node at (3.25, 2.5){$\times$};
      % \node at (5.75, 1){$\times$};
        
    \end{tikzpicture}
    }
\end{equation}
where columns of $\mathbf{\Phi}$ indicate the POD modes. By choosing $p$-significant modes, the branch network aims to replicate the diagonal scaling ($\mathbf{\Sigma} \mathbf{\mathbf{V}^{*}}$). The determination of the effective number of modes is a hyper-parameter that requires careful consideration.
% For accurate reconstruction of the displacement vector field, specifically in this case, utilizing just $16$ modes proves to be sufficient. 
In the proposed POD-DeepONet architecture, the strain tensor $\vepsilon(\vx, \macStrain)$ is obtained by projecting the kinematic relation on the first $p$ modes of the POD, analogous to Galerkin methods
\begin{equation}
    \label{eq:kinematicPOD}
    \vepsilon(\vx, \macStrain) \approx \dfrac{1}{2}\ \sum_{k=1}^{p} b_k(\macStrain)\, (\nabla \phi_k(\vx) + \nabla \phi_k^T(\vx)) + \dfrac{1}{2} (\nabla \phi_0(\vx) + \nabla \phi_0^T(\vx)),
\end{equation}
where $\{\phi_1, \phi_2, \cdots, \phi_p\}$ are the POD basis of $\vu(\vx, \macStrain)$ and $\phi_0$ is the mean function. Further, the microscale stress tensor $\vsigma(\vx, \macStrain)$ is computed by embedding the constitutive relations into the model.

%%%%%%%%%%%%%%%%%%%%%%%%%%%%%%
\section{Numerical experiments}
In this section, we explore the proposed hybrid multiscale method for simulating two standard test cases in computational solid mechanics: L-profile and Cook's membrane. 

\subsection{Problem setup}
\label{sec:problem_setup}

For a comprehensive understanding of the experimental setup, including detailed specifics of the test cases, we refer to \cite{mca28040091}. In this work, we restrict ourselves to two-dimensional test cases, where a plane strain case and small strains are always assumed. The RVE under consideration, which represents the heterogeneous microstructure, is chosen as a commonly applied geometry in the mechanical analysis of composite materials with a fiber volume fraction of 55\%, see \Cref{fig:fedeeponet}(b). The fibers are assumed to behave linearly elastic,
\begin{equation}
    \label{eq:linElas}
    \vsigma(\vx,\macStrain) = K_f \, \tr(\vepsilon(\vx, \macStrain))\sI + G_f \, \vepsilon^{\textrm{D}}(\vx, \macStrain), \quad \vx \in \Omega^f,
\end{equation}
with bulk modulus $K_f = \SI{4.35e04}{\N\per\mm\squared}$ and shear modulus $G_f = \SI{2.99e04}{\N\per\mm\squared}$, where $\Omega^f \subset \Omega$ represents the domain occupied by the fiber material. The matrix material is modeled with a nonlinear elastic constitutive relationship,
\begin{equation}
    \label{eq:nonlinElas}
    \vsigma(\vx,\macStrain) = K_m \, \tr(\vepsilon(\vx, \macStrain))\sI + G_m(\vepsilon^{\textrm{D}}(\vx, \macStrain))\,\vepsilon^{\textrm{D}}(\vx, \macStrain), \quad \vx \in \Omega^m,
\end{equation}
for any point $\vx$ in the domain occupied by the matrix material $\Omega^m \subset \Omega$. The nonlinearity is induced by the strain-dependent shear modulus
\begin{equation}
    G_m(\vepsilon^{\textrm{D}}(\vx, \macStrain)) = \frac{\alpha_1}{\alpha_2 + \vert\vert\bm{\varepsilon}^{\textrm{D}}(\vx, \macStrain)\vert\vert}.
\end{equation}
Note that the dependence of the bulk modulus $G_m$ on the macroscale strain $\macStrain$ arises from the relation between micro- and macroscale strains within the proposed framework. Here, $\sI$ represents the second-order identity tensor, $\tr \bm{a} = a_{ii}$ is the trace operator, and $\bm{a}^D = \bm{a} - (\tr \bm{a})/3 \sI$ the deviatoric part of a tensor. The material parameters of the nonlinear elastic material model are bulk modulus $K_m = \SI{4.78e03}{\N\per\mm\squared}$ and the scalar parameters $\alpha_1 = \SI{50}{\N\per\mm\squared}$ and $\alpha_2 = \num{0.06}$. The microscale stress tensor $\vsigma(\vx,\macStrain)$ is numerically homogenized to obtain the macroscale stress tensor $\macStress$. Moreover, since the proposed mapping from $\macStrain$ to $\macStress$ is differentiable, the consistent tangent matrix $\mC$ can be obtained using automatic differentiation.

\subsection{Training}
\begin{figure}[ht]
    \centering
    \includegraphics[width=0.95\textwidth]{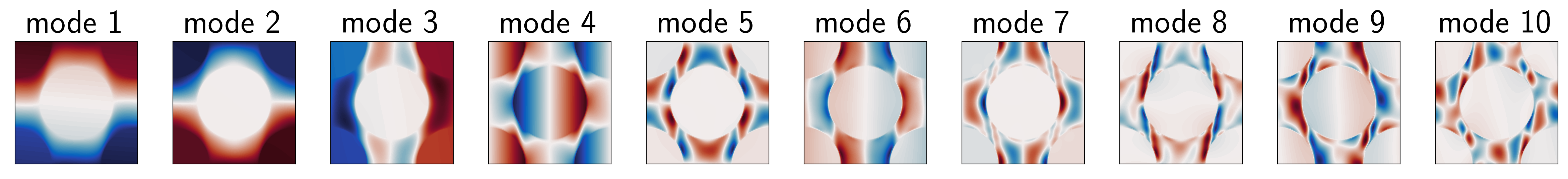}
    \caption{The first 10 POD basis functions for displacement in $y$ direction.}
    \label{fig:modes}
\end{figure}

As mentioned afore, the number of POD modes represents a hyper-parameter. For an accurate reconstruction of the displacement vector field of the specific RVE in this study, utilizing just 16 modes proves to be sufficient. The first 10 POD basis functions of the displacement in $y$-direction are illustrated in \Cref{fig:modes}.
The POD-DeepONet model is trained using data generated from RVE simulations. These computations involve applying various prescribed macroscale strains, which are selected using the Latin hypercube sampling (LHS) technique \cite{lhs_method}. Here, we restrict ourselves to a magnitude of \SI{0.04}{\percent} in tension and compression for each strain component (remember that two-dimensional cases with plane strain assumption are considered). In the branch network, we adopt an architecture comprising $4$ hidden layers, each consisting of $64$ neurons and utilizing the swish activation function. The training process involves 10 batches. We apply a scheduled exponential decay function for the learning rate (initial rate of $0.001$, decay step of $1000$, and decay rate of $0.2$). The loss function employed is the mean-squared error (MSE), optimized using the Adam optimizer. The dataset consists of 1,000 samples and we employ 20\% of the dataset for validation. The surrogate POD-DeepONet model is trained on an NVIDIA GeForce RTX 3090 with CUDA  11.8. After training, the \FEsq simulations are performed on a 12th Gen Intel\textregistered\ Core\texttrademark\ i9-12900K @ 5.20GHz CPU with 48 threads.

\subsection{RVE simulations}
\begin{figure}[h]
    \centering
    \includegraphics[width=0.8\textwidth]{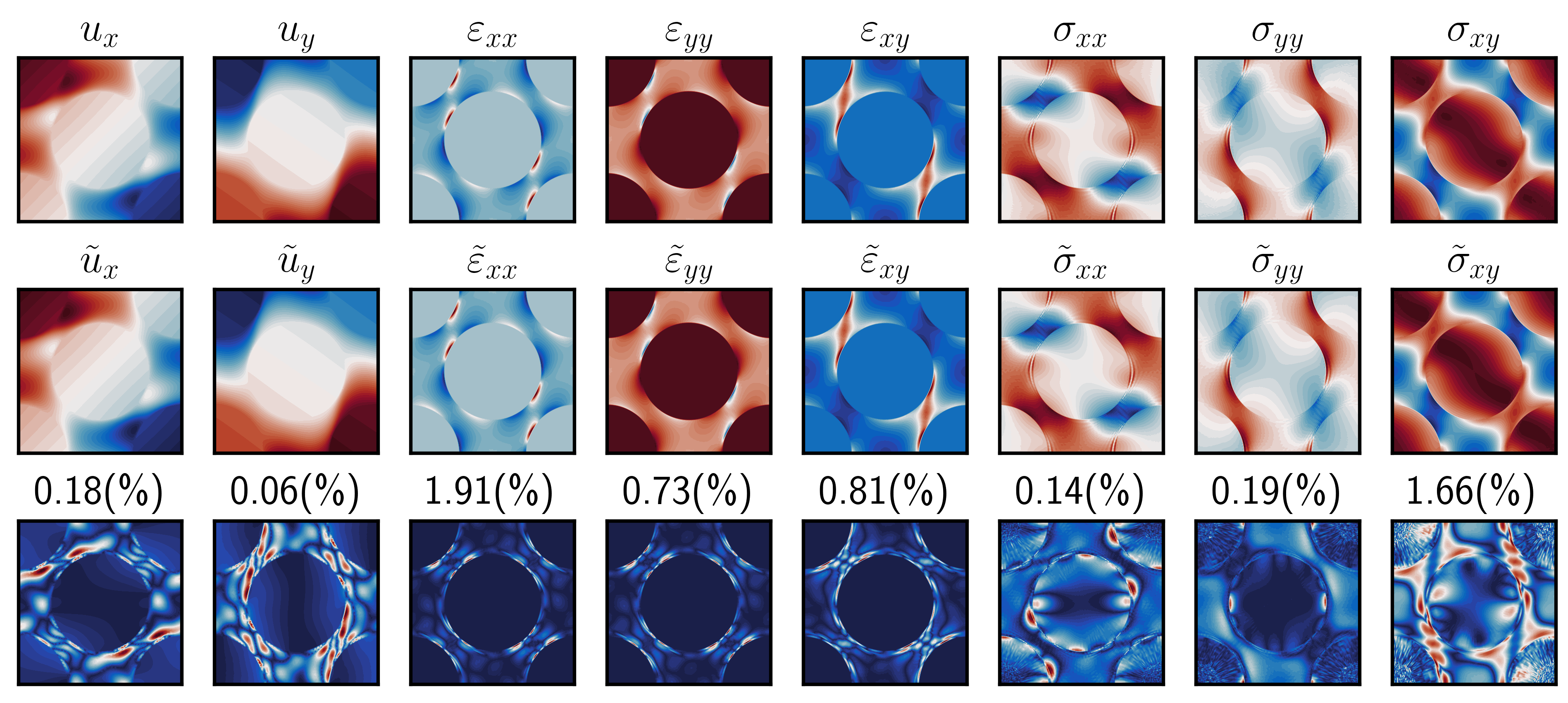}
    \caption{POD-DeepONet simulation results of the microscale RVE for a test sample of the prescribed macroscale strains $\macStrain = \lbrace -0.011, \, -0.036, \, 0.017 \rbrace^\top$. The first row shows reference data obtained from the FE simulation. The second row depicts the predicted solution. Errors are reported in the third row, where contours show the absolute error and titles show the relative $\ell_2$-norm of errors.}
    \label{fig:rve_results}
\end{figure}

We first evaluate the performance of the operator network in solving the microscale boundary-value problem for different prescribed macroscale strains as boundary conditions. \Cref{fig:rve_results} 
illustrates the POD-DeepONet results of the microscale RVE, indicated by $\tilde{\bullet}\,$, for a test sample of the prescribed macroscale strains against reference data obtained from FE simulation. A relative $\ell_2$-norm of errors of less than 2\% is obtained for all the quantities of interest, which are here displacement vector $\vu(\vx, \macStrain)$, microscale strain tensor $\vepsilon(\vx, \macStrain)$ and microscale stress tensor $\vsigma(\vx, \macStrain)$. The hybrid-model is trained on a dataset with 1,000 samples. We observed that POD-DeepONet is able to provide accurate solutions for the microscale boundary-value problem when applying unseen boundary conditions. In the next step, we integrate the trained operator network for multiscale simulations where the macroscale is simulated via classical Bubnov-Galerkin finite elements and the microscale is evaluated by the operator network.

\subsection{Multiscale simulations}
We showcase results concerning simulation accuracy and numerical simulation time for the macroscale geometries considered in this work. The relative $\ell_2$-norm of errors is employed to evaluate the predictive accuracy of the proposed approach compared to the conventional \FEsq method.~\Cref{tab:results} 
\begin{table}[h]
    \centering
    \caption{Relative $\ell_2$-norm of errors and speedup obtained from the proposed hybrid simulation for L-profile and Cook's membrane.}
    \label{tab:results}
        \begin{tabular}{@{}cccccccc@{}}
        \hline
        \rule{0pt}{3ex}Test case & No. of Samples & $\evd_x$ & $\evd_y$ & $\sigma_{xx}$ & $\sigma_{yy}$ & $\tau_{xy}$ & Speed-up \vspace{2pt}\\
        \hline
        \rule{0pt}{2.5ex}L-profile & 1,000 & 0.05\% & 0.01\% & 1.65\% & 1.49\% & 1.46\% & 328 \\
        \rule{0pt}{2.5ex}Cook's & 1,000 & 0.54\% & 0.20\% &  2.07\% & 3.02\% & 1.60\% & 156 \\
        \hline
        \end{tabular}
\end{table}
summarizes the results for L-profile and Cook's membrane test cases for the components of the macroscale displacement vector $\vd = \lbrace \evd_x, \, \evd_y \rbrace^\top$ and the components of the macroscale stress tensor $\macStress = \lbrace \sigma_{xx}, \, \sigma_{yy}, \, \tau_{xy} \rbrace^\top$. A relative $\ell_2$-norm of errors of less than 0.6\% is obtained for the macroscale displacement vector, indicating the applicability of the proposed method for multiscale simulations. The highest errors are related to the normal stresses and are equal to 1.65\% for the L-profile and 3.02\% for the Cook's membrane, respectively. For both test cases, we obtained two orders of magnitude speed-up compared to the reference \FEsq simulation, which is obtained by drawing on state-of-the-art high-performance computing libraries and just-in-time compilation, see \cite{Trger2023} for details on the implementation. These results motivate the application of operator learning for multiscale problems and the development of efficient and accurate surrogate models of microscale physical systems. 
In contrast to conventional methods where the microscale is completely substituted by a surrogate model, our method utilizes operator networks for approximating the solution operator of the microscale. The findings presented show that our method maintains the same level of prediction accuracy while requiring significantly less data to train the surrogate model. However, it should be noted that this approach slightly reduces the speed-up advantage generally seen in existing methods, due to the need for predicting microscale quantities and performing homogenization.

\section{Conclusions}
In this contribution, we utilize the POD-DeepONet framework and formulate a constitutive relations-aware operator network to approximate the microscale solution quantities and enhance multiscale \FEsq simulations, achieving significant reductions in computational overhead while maintaining accuracy in two specific test scenarios.
Our research introduces the first surrogate model based on operator learning that offers microscale solutions for the heterogeneous microstructure. Unlike most studies that focus on global mappings, our model integrates known microscale physical relationships, i.e. the kinematic and constitutive relations.
This integration enables precise capture of microscale variations, enhancing the accuracy of macroscale computations and yielding improved overall outcomes. First, we demonstrate the performance of the operator network in solving the microscale boundary-value problem when applying unseen boundary conditions. In this case, we obtain a relative $\ell_2$-norm of errors of less than 2\% for all the quantities of interest when the model is trained using a dataset with only 1,000 samples. Further, we incorporate the trained operator network into a multiscale solver where the global macroscale mechanics are simulated using FEM and the microscale quantities are evaluated by the operator network. Simulation results are reported for two canonical test cases, where a relative $\ell_2$-norm of errors of less than 1\% is obtained for the macroscale displacement vector. Additionally, our implementation results in a speedup of up to 328 times compared to the reference \FEsq simulation. 

For further improvement, the utilization of physics-informed neural networks for continuum micro-mechanics \cite{HENKES2022114790} can enhance our approach. By incorporating PINNs, boundary conditions and the residual of the PDE, i.e. the balance of linear momentum, can be imposed as additional parameters to the loss function, thereby integrating more physics into the training process. This strategy enables the network to learn and enforce physical laws directly, leading to more accurate and robust simulations and will be followed in future works.

\section*{Acknowledgement}
    Hamidreza Eivazi's research was conducted within the Research Training Group CircularLIB, supported by the Ministry of Science and Culture of Lower Saxony with funds from the program zukunft.niedersachsen of the Volkswagen Foundation.
%%%%%%%%%%%%%%%%%%%%%%%%%%%%%%
\bibliographystyle{ieeetr}
\bibliography{pamm-tpl}

\end{document}